\documentclass[11pt]{article}
\usepackage{amssymb}
\usepackage{graphicx}
\usepackage{amsfonts}
\usepackage{amsmath}
\usepackage{longtable,lscape}
\usepackage{amscd}
\usepackage{mathrsfs}
\usepackage{latexsym}
\usepackage{bbm}
\usepackage{float}
\usepackage{url}

\newcommand{\captionfonts}{\footnotesize}
\makeatletter  
\long\def\@makecaption#1#2{%
  \vskip\abovecaptionskip
  \sbox\@tempboxa{{\captionfonts #1: #2}}%
  \ifdim \wd\@tempboxa >\hsize
    {\captionfonts #1: #2\par}
  \else
    \hbox to\hsize{\hfil\box\@tempboxa\hfil}%
  \fi
  \vskip\belowcaptionskip}
\makeatother 

\title{\bf Representing Attitudes Towards Ambiguity in Hilbert Space: Foundations and Applications}
\author{Sandro Sozzo\footnote{This work was supported by QUARTZ (Quantum Information Access and Retrieval Theory), the Marie Sk{\l}odowska-Curie Innovative Training Network 721321 of the European Union's Horizon 2020 research and innovation programme.} \\ School of Business and Centre IQSCS \\ University Road LE1 7RH, 
Leicester (United Kingdom) \\ Email address: ss831@le.ac.uk}

\date{}

\begin{document}

\maketitle

\begin{abstract}
\noindent
We provide here a general mathematical framework to model attitudes towards ambiguity which uses the formalism of quantum theory as a ``purely mathematical formalism, detached from any physical interpretation''. We show that the quantum-theoretic framework enables modelling of the {\it Ellsberg paradox}, but it also successfully applies to more concrete human decision-making (DM) tests involving financial, managerial and medical decisions. In particular, we elaborate a mathematical representation of various empirical studies which reveal that attitudes of managers towards uncertainty shift from {\it ambiguity seeking} to {\it ambiguity aversion}, and viceversa, thus exhibiting {\it hope effects} and {\it fear effects}. The present framework provides a promising direction towards the development of a unified theory of decisions in the presence of uncertainty.
\end{abstract}

\section{Introduction}\label{intro}
Notwithstanding the remarkable success of expected utility theory (EUT), a number of fundamental 
problems need to be solved before a single theory is unanimously accepted to represent human decision-making (DM) under uncertainty.

A first issue is specifying what one means by ``uncertainty''. Following Knight \cite{k1921}, two kinds of uncertainty are possibly present: {\it objective uncertainty}, or {\it risk}, designates situations where probabilities are known or knowable, i.e. can be estimated from past data or calculated by means of mathematical rules. By contrast, {\it subjective uncertainty}, or {\it ambiguity}, designates situations where probabilities are neither known, nor can they be objectively derived, calculated or estimated \cite{gps2008}.

The predominant model of DM under risk was elaborated by von Neumann and Morgenstern \cite{vnm1944}, who identified the axioms allowing to uniquely represent human preferences over {\it lotteries} by maximization of the expected utility (EU) functional. On the one side, however, decision puzzles like the {\it Allais paradox} reveal that some of these axioms are violated in concrete choices \cite{a1953}. And, on the other side, this formulation does not account for ambiguity, which is prevalent in social science and likely influences social science decisions. The {\it Bayesian paradigm} tries to fill this gap introducing the notion of {\it subjective probability}: when the probabilities are not known, people may still form their own {\it beliefs}, or {\it priors}, which generally vary from one person to another. People then update their beliefs according to the Bayes rule of standard probability theory, i.e. the one axiomatized by Kolmogorov ({\it Kolmogorovian probability}) \cite{k1933}. 

In 1950s, Savage extended von Neumann-Morgenstern EUT in agreement with the Bayesian paradigm, presenting a set of axioms which, once satisfied, ``compel'' decision-makers to behave as if they had
a {\it single} Kolmogorovian probability with respect to which they maximize EU \cite{s1954}. Savage's formulation, also known as {\it subjective EUT}, provides the foundations of ``rational behaviour'', that is, subjective EUT prescribes how people should behave in the presence of uncertainty, and it has been widely used in decision theory, economics and finance, because of its mathematical simplicity and predictive success. In addition, subjective EUT can be empirically tested. 

Regarding the latter, Ellsberg showed in two simple thought experiments, the {\it two-urn example} and the {\it three-colour example}, that decision-makers do not always maximize EU but, rather, they prefer {\it risky acts} over {\it ambiguous acts}, a behaviour known as {\it ambiguity aversion} \cite{e1961}. {\it Ellsberg preferences} particularly violate the famous {\it sure thing principle}, one of the building blocks of Savage's axiomatization, and they have been empirically confirmed several times (see, e.g., the reviews in \cite{ejt2012} and \cite{ms2014}). These well-documented violations of EUT have led many scholars to elaborate alternative DM models, which even include representation of beliefs by more general structures than a single Kolmogorovian probability measure (see, e.g., the reviews in \cite{ejt2012}, \cite{ms2014} and \cite{gm2013}).

Things are not cleared out if one considers more concrete DM tests, in which financial, managerial, medical, etc., decisions are taken in the presence of uncertainty. In these cases, indeed, different {\it attitudes towards ambiguity}, e.g., {\it ambiguity seeking}, arise in addition to the ambiguity aversion revealed by Ellsberg preferences. To realize the possibility of attitude reversal, consider the following example. 

Suppose your doctor tells you that there is a certain probability that you have a serious disease. You then look for alternative opinions. Some doctors estimate that the probability is much lower than the one originally estimated, while others estimate that the probability is much higher. Which option would you ``prefer'' -- the former which is risky, or the latter which is ambiguous? Intuition suggests that the degree of probability will play a fundamental role in the final decision. In fact, if the probability is low, then it is reasonable to assume that a fear effect occurs and you prefer the risky option, thus showing an ambiguity aversion behaviour. On the other side, if the probability is high, then it is reasonable to assume that a hope effect occurs and you instead prefer the ambiguous option, thus showing an ambiguity seeking behaviour \cite{vc1999}.

A comparison between a risky and an ambiguous option, like the one presented in the two-urn example, was part of two experimental studies on managers, one on DM under environmental uncertainty performed by Viscusi and Chesson \cite{vc1999}, and the other on investment decisions under performance uncertainty  by Ho, Keller and Keltyka \cite{hkk2002}. In both cases, a shift from hope to fear effects, and viceversa, was observed which is incompatible with a theory of rational preferences.

The present study fits an emergent research programme which applies the mathematical formalism of quantum theory to model complex cognitive phenomena, including categorization, decision, judgement, language and perception, where classical Bayesian, or Kolmogorovian, modelling techniques are problematical (see, e.g., \cite{a2009,bb2012,abgs2013,ags2013,hk2013,pb2013}). This research has recently been extended to economics and finance (see, e.g., \cite{h2002,b2004,w2015,hkms2018,hk2018,kp2019,t2020}). In particular, we have recently worked out a theoretical framework to represent individuals' preferences and choices under uncertainty (risk, ambiguity) that uses the  formalism of quantum theory as a ``pure mathematical formalism consisting in non-distributive probability measures and linear vector spaces over complex numbers''. Indeed, this {\it quantum mathematics in Hilbert spaces} has some advantages in modelling the information uncertainty that is induced by a non-controllable context, like a cognitive one \cite{ast2014,as2016,ahs2017,agms2018,s2019a,s2019b}. It must be noted, in this regard, that the present theoretical framework also accords with other attempts to represent human decisions in Hilbert space (see, e.g., \cite{lm2009,ys2010,gh2012,and2017,bkpak2018,dlmv2018,ep2018}).

A new theoretical element of the present quantum-theoretic framework is the introduction of the {\it state of the conceptual entity under investigation} ({\it DM entity}), which has a conceptual, rather than a physical, nature and can change under the interaction with the decision-maker \cite{asdbs2016}. The notion of conceptual state, its representation and connections with subjective probabilities through quantum probability, provide the theoretical tools that enable capturing ambiguity and individual attitudes towards ambiguity as context-induced state changes. The quantum-theoretic framework has been applied to the Ellsberg paradox and recent variants, as recognised by some of the proponents (see, e.g., \cite{m2014}, p. 3836). In this approach, subjective probabilities are represented by quantum probabilities, rather than classical Kolmogorovian probabilities -- structurally, quantum probability is more general than Kolmogorovian probability, as the latter rests on a distributive algebra, while the former does not  \cite{ahs2017}.

In the present paper, we extend our previous findings, showing that the quantum-theoretic framework enables modelling of the Ellsberg two-urn example, also providing a faithful mathematical representation of data collected on a two-urn DM test performed by one the authors, together with other authors \cite{agms2018}. This enables us to reproduce the ambiguity aversion pattern observed in the data. However, the quantum-theoretic framework can also incorporate different attitudes towards ambiguity, including ambiguity seeking behaviour, as well as shifts from one attitude to another. And, indeed, we show here that hope and fear effects in investment choices can be naturally incorporated into the quantum modelling of a DM scenario where a risky option is opposed to an ambiguous option, like in the DM tests in \cite{vc1999} and \cite{hkk2002}.

The results above support a systematic application of quantum structures in economics and the development of a unitary quantum-based theory of DM under uncertainty.

For the sake of completeness, we summarize the content of this paper in the following.

In Sec. \ref{mainstream}, we present the essential  mathematics that is needed to introduce subjective EUT (Sec. \ref{SEUT}), together with the Ellsberg paradox  in the two-urn example (Sec. \ref{ellsberg}) and analyse the empirical study of Ho, Keller and Keltyka \cite{hkk2002} on attitudes reversal in the presence of ambiguity (Sec. \ref{ambiguityattitudes}). We then review in Sec. \ref{QEUT} the quantum-theoretic framework that uses the mathematics of Hilbert space, whose essentials are summarized in Appendix A. Successively, we apply in Sec. \ref{quantumattitudes} the quantum-theoretic framework, model the two-urn example (Sec. \ref{quantumellsberg}) and represent data collected in the two-urn experiment performed by ourselves and revealing an ambiguity aversion pattern (\ref{twourndata}). Finally, we elaborate in Sec. \ref{quantumhopefear} a mathematical model in Hilbert space which accounts for the shifts between ambiguity aversion and ambiguity seeking behaviour in either direction. This allows us to represent hope and fear effects within a single theoretical framework. The mathematical framework works in both investment gain (Sec. \ref{gain}) and loss (Sec. \ref{loss}) scenarios of the empirical study in \cite{hkk2002}, whose data are also represented here (Sec. \ref{managerialdata}). Final comments in Sec. \ref{conclusions} conclude the paper.

\section{Expected utility, paradoxes, ambiguity and its effects\label{mainstream}}
We present in the following sections the essential definitions and results within subjective EUT, together with the Ellsberg paradox and some DM tests revealing individuals' behaviour in the presence of uncertainty. The reader who is interested to deepen these results can refer to \cite{s1954}, \cite{ms2014}, \cite{gm2013}.

The starting point which we will assume in the following is that human preferences are revealed by the decisions of individual agents (or, {\it decision-makers}).

\subsection{Basic mathematical framework of subjective expected utility theory\label{SEUT}}
The first axiomatization of DM under uncertainty was formulated by von Neumann and Morgenstern, who presented a set of axioms allowing to uniquely represent human decisions by means of the maximization of the EU functional with respect to a single Kolmogorovian probability measure \cite{vnm1944}. 

A major limitation of von Neumann-Morgenstern's formulation is that it only deals with the uncertainty that can be formalized by known probabilities ({\it objective uncertainty}, or {\it risk}). On the other hand, situations frequently occur where uncertainty cannot be formalized by known probabilities ({\it subjective uncertainty}, or {\it ambiguity}) \cite{k1921}. As mentioned in Sec. \ref{intro}, a Bayesian approach
would introduce the notion of subjective probability, thus minimizing the distinction  between objective and subjective uncertainty. In a Bayesian approach, 
if probabilities are not known, people anyway form their own {\it beliefs} (or {\it priors}), which may differ across individuals but are still formalized by Kolmogorovian probabilities \cite{gps2008}. As a matter of fact, Savage presented an axiomatic formulation of subjective EUT which extends von Neumann-Morgenstern's in agreement with a Bayesian approach \cite{s1954}.

To present the mathematics of subjective EUT, we preliminarily introduce the following symbols.

Let ${\mathscr S}$ be the set of all {\it physical states of nature} and let ${\mathscr P}({\mathscr S})$ be the power set of ${\mathscr S}$, that is, the set of all subsets of ${\mathscr S}$. Let ${\mathscr A}\subseteq {\mathscr P}({\mathscr S})$ be a $\sigma$-algebra. An element $E \in {\mathscr A}$ denotes an {\it event} in the usual sense. Let $p:{\mathscr A}\subseteq {\mathscr P}({\mathscr S})\longrightarrow [0,1]$ be a {\it Kolmogorovian probability measure} over $\mathscr A$, that is, a normalized countably additive measure satisfying the axioms of Kolmogorov \cite{k1933}.

Then, let ${\mathscr X}$ be the set of all {\it consequences}, and let the function $f: {\mathscr S} \longrightarrow {\mathscr X}$ denote an {\it act}. We denote the set of all acts by $\mathscr{F}$. Moreover, let $\succsim$ denote a {\it weak preference} relation, that is, a relation over the Cartesian product $\mathscr{F} \times \mathscr{F}$ which is complete and transitive. We adopt the usual interpretation of $\succ$ and $\sim$ as {\it strong preference} and {\it indifference} relations, respectively, namely, if a person strongly, or strictly, prefers act $f$ to act $g$, we write $f \succ g$; analogously, if a person is indifferent between $f$ and $g$, we write $f \sim g$.

Next, let $\Re$ be the real line and $u: {\mathscr X} \longrightarrow \Re$ be a {\it utility function}, which we assume to be a strictly increasing and continuous function, as it is usual in the literature.

Let us now introduce some simplifying assumptions, which however do not affect our conclusions in this and the following sections. Firstly, we suppose that the set ${\mathscr S}$ is discrete and finite. Secondly, we suppose that an element $x \in {\mathscr X}$ denotes a monetary payoff, so that ${\mathscr X}\subseteq \Re$.

Let $E_1, E_2, \ldots, E_n$ denote mutually exclusive and exhaustive elementary events, which thus form a partition of ${\mathscr S}$. If $x_i$ is the utility associated by the act $f$ to the event $E_i$, $i \in \{ 1, \ldots, n\}$, then $f$ can be equivalently represented by the 2n-tuple $f=(E_1,x_1;\ldots;E_n,x_n)$, which is interpreted in the usual way as ``we get the payoff $x_1$ if the event $E_1$ occurs, the payoff $x_2$ if the event $E_2$ occurs, \ldots, the payoff $x_n$ if the event $E_n$ occurs.

We finally define the {\it EU functional} $W(f)=\sum_{i=1}^{n}p(E_i)u(x_i)$ associated with the act $f=(E_1,x_1;\ldots;E_n,x_n)$ with respect to the Kolmogorovian probability measure $p$. 

{\it Representation theorem}. If the algebraic structure $({\mathscr F}, \succsim)$ satisfies the axioms of ordinal event independence, comparative probability, non-degeneracy, small event continuity, dominance and the sure thing principle, then, for every $f,g\in \mathscr F$, a {\it unique} Kolmogorovian probability measure $p:{\mathscr A}\subseteq {\mathscr P}({\mathscr S})\longrightarrow [0,1]$ and a {\it unique} (up to positive affine transformations) utility function $u: {\mathscr X} \longrightarrow \Re$ exist such that $f$ is preferred to $g$. In symbols, $f \succsim g$ if and only if the EU of $f$ is greater than the EU of $g$, i.e. $W(f) \ge W(g)$. For every $i\in \{ 1, \ldots, n\}$, the utility value $u(x_i)$ depends on the individual's attitudes towards risk, while $p(E_i)$ is interpreted as a subjective probability, expressing the individual's belief that the event $E_i$ occurs \cite{s1954}.

Savage's representation theorem is at the same time compelling at a {\it normative level} and testable at a {\it descriptive level}. Indeed, regarding the former, if the axioms are intuitively reasonable and decision-makers agree with them, then they {\it must} all behave as if they maximized EU with respect to a single probability measure satisfying Kolmogorov's axioms; furthermore, regarding the latter, the axioms suggest to perform concrete DM tests to confirm/disprove the general validity of subjective EUT, hence of the axioms themselves. This is why Savage's EU formulation is typically accepted to prescribe ``how rational agents should behave in the presence of uncertainty'' providing, in this way, the decision-theoretic foundation of the Bayesian paradigm. However, on the one side, the theory offers very little about where beliefs come from and how they should be calculated while, on the other side, DM tests, performed since the 1960s, have systematically found deviations from that rational behaviour in concrete situations. This will be the content of Secs. \ref{ellsberg} and \ref{ambiguityattitudes}.

\subsection{The Ellsberg paradox\label{ellsberg}}
In 1961, Daniel Ellsberg presented various thought experiments in which he suggested that decision-makers would prefer acts with known (or objective) probabilities over acts with unknown (or subjective) probabilities, regardless of EU maximization. In other words, they would prefer probabilized to non-probabilized uncertainty \cite{e1961}. Ellsberg did not perform the experiments himself, but two of these have meanwhile became famous, namely, the {\it three-color example} and the {\it two-urn example}.

We discuss here the two-urn example, because it is the paradigmatic example that is used in more concrete DM tests on ambiguity and ambiguity attitudes.

Consider two urns, {\it urn I} with 100 balls that are either red or black in unknown proportion, and {\it urn II} exactly with 50 red balls and 50 black balls. One ball is to be drawn at random from each urn. Then, free of charge, a person is asked to bet on pairs of the acts $f_1$, $f_2$, $f_3$ and $f_4$ in Table 1.

\noindent 
\begin{table} \label{table01}
\begin{center}
\begin{tabular}{|p{0.7cm}|p{2.3cm}|p{2.3cm}||p{2.3cm}|p{2.3cm}|}
\hline
\multicolumn{1}{|c|}{} & \multicolumn{2}{c||}{Urn I} & \multicolumn{2}{c|}{Urn II} \\
\hline
 Acts & $E_R$: red ball & $E_B$: black ball & $E_R$: red ball & $E_B$: black ball \\ 
\multicolumn{1}{|c|}{} & \multicolumn{1}{c|}{$p_{R} \in [0,1]$}  & \multicolumn{1}{c||}{$p_{B}=1-p_{R}$} &  \multicolumn{1}{c|}{$p_R=1/2$} & \multicolumn{1}{c|}{$p_B=1/2$}  \\

\hline
$f_1$ & \$100 & \$0 &  &  \\ 
\hline
$f_2$ &  &  & \$100 & \$0 \\ 
\hline
$f_3$ & \$0 & \$100 &  &  \\ 
\hline
$f_4$ &  &  & \$0 & \$100 \\ 
\hline
\end{tabular}
\end{center}
{\bf Table 1.} Representation of events, payoffs and acts in the Ellsberg two-urn example.
\end{table}
We denote 
  the event ``a red ball is drawn'' and ``a black ball is drawn'' by $E_R$ and $E_B$, respectively. Then, we observe that $f_2$ and $f_4$ are {\it unambiguous acts}, because they are associated with events over known probabilities, 0.5 in this case, whereas $f_1$ and $f_3$ are {\it ambiguous acts}, because they are associated with events over unknown probabilities, ranging from 0 to 1 in this case. This distinction led  
  Ellsberg to suggest that, while decision-makers will be generally indifferent between acts $f_1$ and $f_3$ and between acts $f_2$ and $f_4$, they will instead generally prefer  $f_2$ over $f_1$ and $f_4$ over $f_3$, a behaviour called {\it ambiguity aversion} by Ellsberg \cite{e1961}.

The predictions of subjective EUT are incompatible with the {\it Ellsberg preferences} $f_2 \succ f_1$ and $f_4 \succ f_3$, hence behaviour of a decision-maker who is psychologically influenced by ambiguity cannot be reproduced by subjective EUT. Indeed, maximization of EU entails consistency of preferences, that is, $f_2 \succ f_1$ if and only if $f_3 \succ f_4$. To show this, suppose that decision-makers assign subjective probabilities $p_R$ and $p_B=1-p_R$ to the events $E_R$ and $E_B$, respectively. Then, an EU maximizer will prefer $f_2$ over $f_1$ if and only if $W(f_2)>W(f_1)$, which is equivalent to the condition $(p_R-\frac{1}{2})(u(100)-u(0))<0$, where $u(0)$ and $u(100)$ denote the utilities associated with the payoffs 0 and 100, respectively. On the contrary, the same EU maximizer will prefer $f_4$ over $f_3$ if and only if $W(f_4)>W(f_3)$, which is equivalent to the condition $(p_R-\frac{1}{2})(u(100)-u(0))>0$. These conditions cannot be simultaneously satisfied, which entails that one cannot find a single Kolmogorovian probability measure $p$ such that $f_2 \succ f_1$ and $f_4 \succ f_3$ by maximization of the EU functional with respect to $p$, whence the {\it Ellsberg paradox}.

Decision tests on Ellsberg urns have been performed since 1960s and they generally confirm Ellsberg preferences against subjective EUT, hence an ambiguity aversion attitude of decision-makers (see, e.g., \cite{ms2014} for a review of experimental studies).

We have recently performed a whole set of DM tests, including the two-urn example, to check the quantum-theoretic framework introduced in Sec. \ref{intro}. In \cite{agms2018}, we presented a sample of 200 participants with a questionnaire in which they had to choose between the pairs of acts ``$f_1$ versus $f_2$'' and ``$f_3$ versus $f_4$'' in Table 1. Respondents were provided with a paper similar to the one in Figure 1.\footnote{For the sake of simplicity, we assumed that each choice concerned two alternatives, hence indifference between acts was not a possible option.}
\begin{figure}
\begin{center}
\includegraphics[scale=0.5]{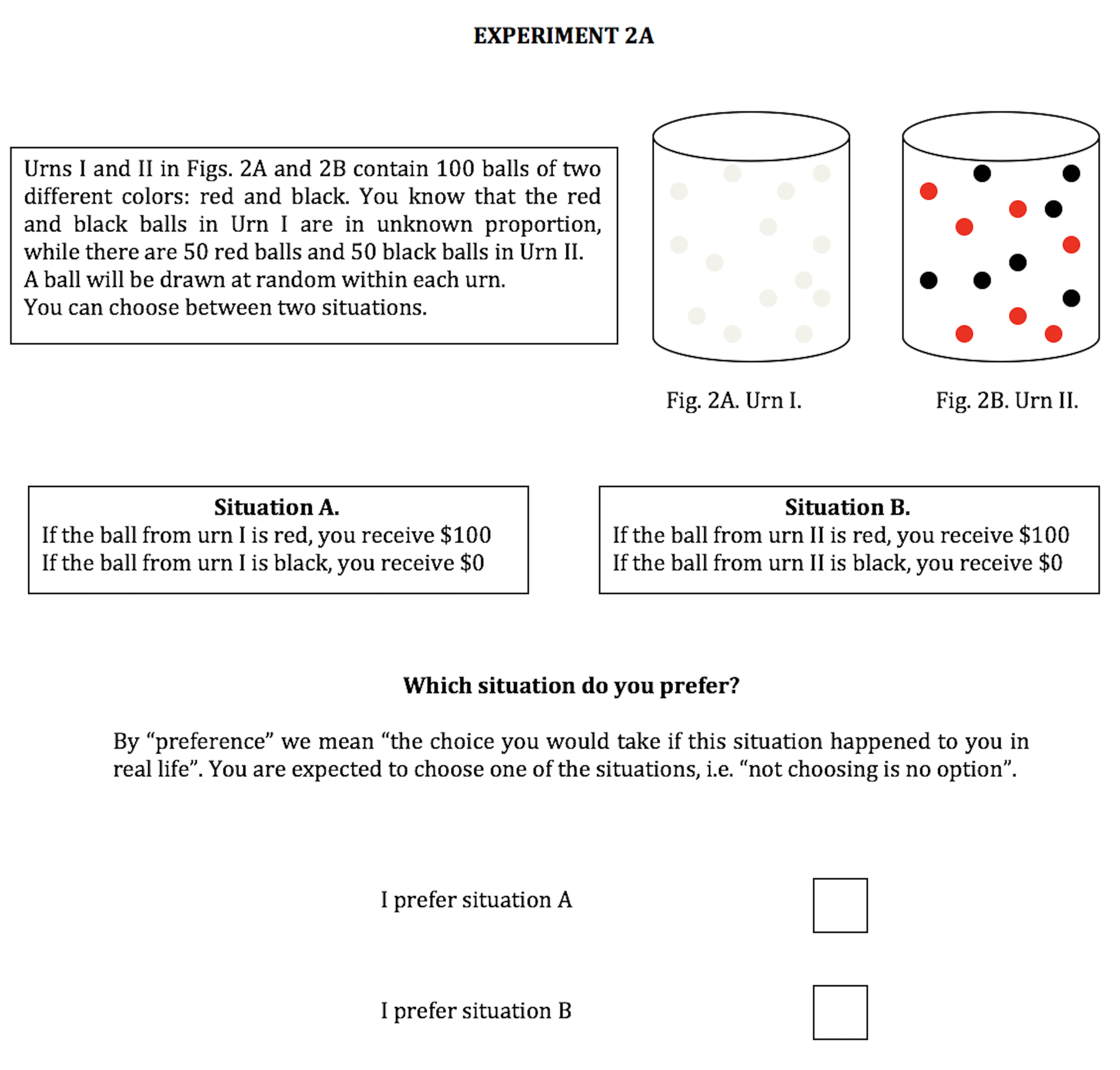}
\end{center}
{\bf Figure 1.} A sample of the questionnaire on the two-urn example. It corresponds to the choice between acts $f_1$ and $f_2$ in Table 1.
\end{figure}
In the two-urn test, 26 participants chose acts $f_1$ and $f_3$, 10 chose $f_1$ and $f_4$, 6 chose $f_2$ and $f_3$, and 158 chose $f_2$ and $f_4$. Hence, overall 164 respondents over 200 preferred act $f_2$ over act $f_1$, for a preference rate of 164/200=0.82 (the difference is significant, $p=1.49E-24$). Moreover, 168 respondents over 200 preferred act $f_4$ over act $f_3$, for a preference rate of 168/200=0.84 (the difference is again significant, $p=1.25E-28$). Finally, 16 respondents over 200 preferred either $f_1$ and $f_4$ or $f_2$ and $f_3$, for an inversion rate of 184/200=0.92. This pattern accords with Ellsberg preferences and straightly points towards ambiguity aversion, thus confirming existing results in empirical literature.

The Ellsberg paradox and other Ellsberg-type puzzles put at stake both the descriptive and normative foundations of subjective EUT, which led various scholars to propose alternatives to subjective EUT, in which more general, including non-Kolmogorovian, mathematical structures are used to represent subjective uncertainty. Major non-EU models include {\it Choquet EU}, {\it cumulative prospect theory}, {\it maxmin EU}, {\it alpha maxmin EU}, {\it smooth preferences}, {\it variational preferences}, etc. (see, e.g., the reviews in \cite{ms2014} and \cite{gm2013}). 

However, Mark Machina elaborated in 2009 two variants of Ellsberg examples, the {\it 50/51 example} and the {\it reflection example}, which challenge major non-EU models in a similar way as Ellsberg examples challenge subjective EUT \cite{m2009,blhp2011}. {\it Machina preferences} have been confirmed in two tests against the predictions of both subjective EUT and its non-EU generalizations  \cite{agms2018,lhp2010}. The implication of Ellsberg and Machina paradoxes is that a unified theoretic approach to represent human preferences and choices under uncertainty is still an unachieved goal \cite{lhp2010}.

\subsection{Empirical studies on ambiguity attitudes\label{ambiguityattitudes}}
We have seen in Sec. \ref{ellsberg} that, in each pair of acts of the two-urn example, people are asked to choose between a {\it risky option}, that is, an option with known probability of getting a given consequence, and an {\it ambiguous option}, that is, an option with unknown probability, but belonging to a given range, of getting the same consequence. This is exactly the experimental setting that is designed to test individual attitudes towards ambiguity in more concrete DM situations, involving medical, managerial and financial decisions.

Tests in scenarios different from urns have revealed different attitudes towards ambiguity, namely, {\it ambiguity neutral} and {\it ambiguity seeking}, in addition to the {\it ambiguity aversion} identified in the two-urn example (see, e.g., \cite{ejt2012}, \cite{ms2014} for a review of these studies). In these cases, attitudes towards ambiguity depend on:

(i) likelihood of uncertain events;

(ii) domain of the consequences;

(iii) source that generates ambiguity \cite{tvdk2015}.

We review some of these studies in the present section. To this end, we preliminarily introduce two notions which express the success or failure of a financial operation, as follows. A {\it gain} is realized when the result of an operation made by the decision-maker is above expectations. A {\it loss} is realized when the result of an operation made by the decision-maker is below expectations.

For example, previous empirical studies on financial decisions in the presence of uncertainty consisted in asking participants to compare a risky option with a given probability $\bar{p}$ to realize a gain (respectively, a loss), with an ambiguous option with a probability to realize a gain (respectively, a loss) ranging between $\bar{p}-\Delta$ and $\bar{p}+\Delta$ (see, e.g., \cite{ejt2012,ms2014,vc1999}). The authors  found that in general (i) plays a fundamental role in determining ambiguity attitudes. Indeed, if the probability of a gain is high, then a {\it fear effect} occurs in which people tend to be ambiguity averse. But, as the probability of a gain decreases, people tend to be less ambiguity averse, reaching a {\it crossover point} in which they become ambiguity seeking, which indicates a shift from a fear to a {\it hope effect}. Viceversa, if the probability of a loss is high, then a  hope effect occurs in which people tend to be ambiguity seeking. But, as the probability of a loss decreases, people tend to be less ambiguity seeking, reaching a crossover point in which they become ambiguity averse, which indicates a shift from a hope to a fear effect.

This empirical pattern was confirmed at high probabilities by a test performed by Ho, Keller and Keltyka \cite{hkk2002} on 40 MBA students. Managers are typically provided with a target, or a benchmark, to measure their performance. Thus, a management decision corresponds to a gain (loss) if it leads to a higher (lower) outcome than the benchmark. The authors considered two experiments in a ``within subjects design'', as follows.

{\it ROI experiment}. In this experiment, the {\it return on investment} (ROI) had a benchmark of 16\%. In the loss scenario, participants had to compare a risky option with a probability of a loss equal to 60\%, with an ambiguous option with a probability of a loss ranging between 40\% and 80\%. In the gain scenario, participants had instead to compare a risky option with a probability of a gain equal to 63\%, with an ambiguous option with a probability of a gain ranging between 42\% and 84\%.

In the ROI experiment, the authors found that, in the loss scenario, 59\% of the participants preferred the ambiguous option, thus revealing an ambiguity seeking attitude, hence the presence of a hope effect (41\% preferred instead the risky option). On the contrary, in the gain scenario, 66\% of the participants preferred the risky option, thus revealing an ambiguity averse attitude, hence the presence of a fear effect. These findings agree with those in previous studies (see, e.g., \cite{ejt2012,ms2014,vc1999}).

{\it IRR experiment}. In this experiment, the {\it internal rate of return} (IRR) had a benchmark of 15\%. In the loss scenario, participants had to compare a risky option with a probability of a loss equal to 65\%, with an ambiguous option with a probability of a loss ranging between 45\% and 85\%. In the gain scenario, participants had instead to compare a risky option with a probability of a gain equal to 68\%, with an ambiguous option with a probability of a gain ranging between 47\% and 89\%.

In the IRR experiment, the authors found that, in the loss scenario, 65\% of the participants preferred the ambiguous option, thus revealing an ambiguity seeking attitude, hence the presence of a hope effect (35\% preferred instead the risky option). On the contrary, in the gain scenario, 62\% of the participants preferred the risky option, thus revealing an ambiguity averse attitude, hence the presence of a fear effect. These findings are consistent with those in the ROI experiment and agree with the patterns in previous studies (see, e.g., \cite{ejt2012,ms2014,vc1999}).

Additional experiments were performed in \cite{hkk2002} in which other sources of uncertainty, as outcome ambiguity, were investigated. However, we will not deal with these additional experiments here, for the sake of brevity. 

\section{A quantum-theoretic framework for human decisions\label{QEUT}}
We present in this section the essentials of the general quantum-theoretic framework we have recently elaborated to model human decisions, which proposes a unitary solution to the puzzles in Secs. \ref{ellsberg} and \ref{ambiguityattitudes} \cite{ast2014,as2016,ahs2017,agms2018,s2019a,s2019b}.

The quantum-theoretic framework needs (some of) the mathematics of Hilbert space, hence, before proceeding further, it is worth to explain why we believe that this mathematical framework is well suited to represent the entities, states, context-induced state changes and subjective probabilities involved in a DM process.

The use of a Hilbert space formalism is suggested by the analogy existing between the description of an experiment in quantum physics and the description of a cognitive, e.g., DM, test. This analogy was discussed in detail in \cite{asdbs2016}, but we present the salient points of it in the following.

An experiment on a quantum entity is typically performed in a laboratory, i.e. a space-temporal domain. The quantum entity preliminarily undergoes a preparation procedure designed by the experimenter, at the end of which ``the entity is in a specific state''. This state expresses the ``physical reality'' of the quantum entity, in the sense that, as a consequence of being in that state, the entity has some ``actual'' properties independently of any measurement that can be performed on it ({\it realistic part}). When a measurement is performed on the quantum entity, the macroscopic apparatus acts as a measurement context which interacts, {\it on a physical level}, with the entity and changes its state in a way that is generally neither controllable nor predictable ({\it operational part}). Then, the quantum entity, its states, contexts, properties, and the mutual statistical relations are canonically represented in the Hilbert space formalism.

A similar {\it realistic-operational description} can be provided of a DM test \cite{asdbs2016}. A DM test is typically performed in a ``laboratory'', i.e. a room or space-temporal domain. Suppose, for example, that a DM test is performed in which a sample of participants have to pick their choices from a list on a questionnaire. The information contained in the questionnaire and the meaning content of the situation that is the object of the decision (literally what is written in the questionnaire and which has to be decided on) define a preparation procedure for a {\it conceptual DM entity} $\Omega_{DM}$, at the end of which we can say that ``$\Omega_{DM}$ is in a defined {\it state} $p_v$''. Thus, a preparation procedure is literally something which does take place when a DM test is performed and each participant ``is confronted with this one and unique state'', independently of any belief the participant has about it, because the state was prepared by the experimenter designing the test, long before and independent of any individual participating in the test. This state has a conceptual, rather than a physical, nature, but is a ``state of affairs'', because it expresses the meaning content of the questionnaire that was prepared by the experimenter. As such, it is not a mental state or a state of belief. In addition, this state is independent of any operation that can be performed on $\Omega_{DM}$, hence it expresses the ``conceptual reality'' of the entity at a given time. Having a conceptual nature, this state must be distinguished from a physical state of nature (see Sec. \ref{SEUT}).\footnote{The notions of ``conceptual entity'' and ``conceptual state'' will be specified in the applications to concrete examples, e.g., the Ellsberg two-urn example (Sec. \ref{quantumellsberg}).} The state of $\Omega_{DM}$ can change under the effect of a {\it context}, which has a cognitive nature. Indeed, when the DM test is performed and a participant is asked to make a choice, the individual acts as a context which interacts, {\it on a cognitive level}, with $\Omega_{DM}$ and changes its state in a way that is generally neither controllable nor predictable. More precisely, when participant number 1 enters the room and fills out the questionnaire, the participant ``interacts with this conceptual state presented to her/him in the test'', but prepared, e.g., on a paper, by the experimenter, and changes this state. When participant number 2 enters the laboratory, the participant again interacts with this independently prepared conceptual state and contextually changes it, etc. We will see that it is this {\it potentiality of the state to change under a context that allows the state to capture aspects of ambiguity and ambiguity aversion}. Indeed, ambiguity aversion or ambiguity seeking states are states that are contextually changed in the course of the DM test. 

The realistic-operational description above suggests representing  canonically conceptual entities, states, contexts, properties and mutual statistical relations in the Hilbert space formalism of quantum theory. Indeed, the Hilbert space formalism proves that this dynamics of contextual change can be modelled in specific cases, and we believe that it can be successfully applied to all other types of DM tests.

It is important to note that the above description of a DM process {\it differs} from that of other quantum-based approaches to DM, where the state corresponds to a {\it mental state of the decision-maker}, which can change under the influence of an external context (see, e.g., \cite{bb2012,hk2013}). While the latter description is frequently used, we think that it is closer to a {\it quantum Bayesianism interpretation} of the Hilbert space formalism of quantum theory (see, e.g., \cite{k2016}) and, additionally, does not completely capture the elements involved in a DM process, preparation, contextual interaction, state change. This is why we prefer to adopt the description of the DM process presented here, as we think that it more completely describes the dynamics of a DM process. In addition, it more closely adheres to the interpretation of the Hilbert space formalism expounded in modern manuals of quantum theory.

We also notice that the notion of ``state of a DM entity'' introduces a new element, not directly related to beliefs and not previously used in cognitive science, at the best of our knowledge. The notion is mainly borrowed from physics, as a test in cognitive science, like an experiment in physics, is a bridge between a preparation and a measurement. We believe, that the notion of state should be a constitutive element of any theory statistically connecting entities, contexts, experiments and dynamics. 

Coming now to the specific DM situations presented in Sec. \ref{mainstream}, let $\Sigma_{DM}$ be the set of all states of $\Omega_{DM}$ and let ${\mathscr E}$ be the set of all events which may occur. For every $p_v\in \Sigma_{DM}$, let $\mu(E,p_v)$ be the (subjective) probability that $E$ occurs when $\Omega_{DM}$ is in the state $p_v$.

Then, let $E_1, E_2, \ldots, E_n \in \mathscr E$ denote mutually exclusive and exhaustive elementary events, let ${\mathscr X}\subseteq \Re$ be the set of all consequences (assumed to denote monetary payoffs), and let, for every $i \in \{ 1,\dots,n\}$, the act $f$ map the elementary event $E_i \in \mathscr E$ into the payoff  $x_i\in \Re$, so that $f=(E_1,x_1;\ldots;E_n,x_n)$. Finally, let $u: {\mathscr X} \longrightarrow \Re$ be a continuous strictly increasing utility function expressing individual attitudes towards risk.

We use the canonical Hilbert space notation and representation reviewed in Appendix A, and associate $\Omega_{DM}$ with a Hilbert space $\mathscr H$ over the field $\mathbb C$ of complex numbers. The number $n$ of mutually exclusive and exhaustive elementary events entails that $\mathscr H$ can be chosen to be isomorphic to the Hilbert space ${\mathbb C}^n$ of all n-tuples of complex numbers. Let $\{ |e_1\rangle, |e_2\rangle, \ldots, |e_n\rangle\}$ be the canonical orthonormal (ON) basis of ${\mathbb C}^n$, where $|e_1\rangle=(1,0,\ldots 0)$, $|e_2\rangle=(0,1,\ldots)$, \ldots, $|e_n\rangle=(0,0,\ldots, 1)$.

A state  $p_v\in \Sigma_{DM}$ of $\Omega_{DM}$ is represented by a unit vector $|v\rangle\in {\mathbb C}^n$, $\langle v|v\rangle=1$. 

An event $E \in \mathscr E$ is represented by an orthogonal projection operator $\hat{P}$ over ${\mathbb C}^n$. The set ${\mathscr L}( {\mathbb C}^n)$ of all orthogonal projection operators over ${\mathbb C}^n$ 
 has a non-distributive lattice structure, unlike a Boolean algebra.

It follows from the above quantum representation of events that, for every $i \in \{1,\ldots, n \}$, the elementary event $E_i$ is represented by the 1-dimensional orthogonal projection operator $\hat{P}_i=|e_i \rangle \langle e_i|$.

For every state $p_v \in \Sigma_{DM}$ of $\Omega_{DM}$, represented by the unit vector $|v\rangle=\sum_{i=1}^{n}\langle \alpha_i|v\rangle |e_i\rangle\in {\mathbb C}^n$, the function
\begin{equation} \label{Bornprobabilitymeasure}
\mu_{v}: \hat{P} \in {\mathscr L}( {\mathbb C}^n) \longmapsto  \mu_{v}(\hat{P})=\langle v | \hat{P}|v\rangle\in [0,1]
\end{equation}
induced by the Born rule, is a quantum probability measure over ${\mathscr L}( {\mathbb C}^n)$. In particular, we identify $\mu_{v}(\hat{P})$ with the (subjective) probability $\mu(E,p_v)$ that the event $E$, represented by the orthogonal projection operator $\hat{P}$, occurs when $\Omega_{DM}$ is in the state $p_v$. Thus, in particular, for every $i\in \{ 1, \ldots, n\}$,
\begin{equation} \label{quantumprobability}
\mu(E_i,p_v)=\langle v|\hat{P}_i|v\rangle=|\langle e_i|v\rangle|^{2}
\end{equation}

Let us now represent acts by using the quantum mathematical formalism. The act $f=(E_1,x_1;\ldots;E_n,x_n)$ is represented by the hermitian operator
\begin{equation} \label{quantumact}
\hat{F}=\sum_{i=1}^{n} u(x_i)\hat{P}_i=\sum_{i=1}^{n} u(x_i)|e_i\rangle\langle e_i|
\end{equation}
Then, we introduce, for every $p_v \in \Sigma_{DM}$,  the functional ``EU in the state $p_v$'' $W_v:{\mathscr F} \longrightarrow \Re$, as follows. For every $f \in {\mathscr F}$,
\begin{eqnarray} \label{quantumexpectedutility}
W_{v}(f)&=&\langle v| \hat{F}| v \rangle=\langle v| \Big ( \sum_{i=1}^{n} u(x_i)\hat{P}_i   \Big ) |v\rangle \nonumber \\
=\sum_{i=1}^{n} u(x_i) \langle v|\hat{P}_i|v\rangle&=&\sum_{i=1}^{n} u(x_i) |\langle e_i|v\rangle|^{2}=\sum_{i=1}^{n} \mu(E_i,p_v)u(x_i)
\end{eqnarray}
where we have used (\ref{quantumprobability}) and (\ref{quantumact}). Equation (\ref{quantumexpectedutility}) generalizes the usual EU formula of Sec. \ref{SEUT}.

We observe that the EU generally depends on the state $p_v$ of the DM entity $\Omega_{DM}$. When $W_v(f)$ does (not) explicitly depend on the state $p_v$, then the act $f$ is (un)ambiguous, in the sense specified in Sec. \ref{ellsberg}. This agrees with the insight above that the state $p_v$ mathematically and conceptually incorporates the presence of ambiguity.

Let us now come to the description of the DM process. The state of the DM entity can change under the effect of a context, which has a cognitive nature, as we have seen above. An example of such a context-dependence is given by an interaction with the decision-maker. Indeed, suppose that, when the decision-maker is presented with a questionnaire involving a choice between the acts $f$ and $g$, the DM entity $\Omega_{DM}$ is in the initial state $p_{v_0}$. As the decision-maker starts comparing $f$ and $g$ and before a decision is made, this mental activity can be described as a context interacting with $\Omega_{DM}$ and changing $p_{v_0}$ into a new state. The type of state change depends on the decision-maker's attitude towards ambiguity. More specifically, a given attitude towards ambiguity, say ambiguity aversion, will determine a given change of state of the DM entity to a state $p_v$, inducing the decision-maker to prefer, say $f$. But, a different attitude towards ambiguity, say ambiguity seeking, will determine a different change of state of the DM entity to a state $p_{w}$, leading the decision-maker to instead prefer $g$. In this way, different attitudes towards ambiguity are formalized by different changes of state of the DM entity hence, through (\ref{quantumprobability}), by different (subjective) probability measures.

In symbols, for every $f,g\in \mathscr F$, states $p_v,p_w\in \Sigma_{DM}$ may exist such that $W_{v}(f)>W_{v}(g)$, whereas $W_{w}(f)<W_{w}(g)$, depending on decision-makers' attitudes towards ambiguity. This suggests introducing a {\it state-dependent preference relation} $\succsim_{v}$ on the set of acts $\mathscr F$, as follows. For every $f,g \in {\mathscr F}$ and $p_{v}\in \Sigma_{DM}$,
$f \succsim_{v} g$ if and only if $W_{v}(f) \ge W_{v}(g)$.

We have recently proved that a context-induced state change may explain the {\it inversion of preferences} observed in the Ellsberg and Machina paradox situations, which can be both modelled in the quantum-theoretic framework  \cite{as2016}. In addition, we have provided a quantum representation of various DM tests, including the three-color example \cite{ast2014,agms2018,s2019b}, the 50/51 example \cite{ast2014,agms2018} and the reflection example \cite{ahs2017,agms2018}. 

The results above are important, in our opinion, because the quantum-theoretic framework provides a unitary solution to several paradoxes and pitfalls of rational decision theory. Furthermore, it allows us to draw the following conciliatory result.

According to subjective EUT, decision makers {\it should} maximize EU with respect to a single Kolmogorovian probability measure. The quantum-theoretic framework above shows that decision makers {\it actually} maximize EU with respect to a non-Kolmogorovian, namely quantum, probability measure.

Regarding the two-urn example in Sec. \ref{ellsberg}, the quantum-theoretic framework can be applied too, as we have proved in \cite{s2019a}. However, due to its paradigmatic character to represent ambiguity attitudes, we want to dedicate a separate section to the modelling.

\section{A quantum model for ambiguity aversion effects\label{quantumattitudes}}
In this section, we particularize the quantum-theoretic framework in Sec. \ref{QEUT} to the two-urn example, proving that it enables representation of the empirical data in Sec. \ref{ellsberg} and that it enables modelling of hope and fear effects in management decision tests.

\subsection{Quantum representation of the two-urn example\label{quantumellsberg}}
The two-urn example defines two conceptual entities, DM entity $\Omega_{DM}^{I}$ which is the urn with 100 red or black balls in unknown proportion, and DM entity $\Omega_{DM}^{II}$ which is the urn with 50 red balls and 50 black balls. 

Let $E_R$ and $E_B$ denote the exhaustive and mutually exclusive elementary events ``a red ball is drawn'' and ``a black ball is drawn'', respectively. Both $\Omega_{DM}^{I}$ and $\Omega_{DM}^{II}$ are thus associated with a 2-dimensional complex Hilbert space, which we choose to be ${\mathbb C}^{2}$. Let $|e_1\rangle=(1,0)$ and $|e_2\rangle=(0,1)$ be the unit vectors of the canonical ON basis of ${\mathbb C}^{2}$. The elementary events $E_R$ and $E_B$ are represented by the projection operators $\hat{P}_{R}=|e_1\rangle\langle e_1|$ and $\hat{P}_{B}=|e_2\rangle\langle e_2|=\mathbbmss{1}-\hat{P}_R$, projecting onto the 1-dimensional subspace generated by $|e_1\rangle$ and $|e_2\rangle$, respectively.

Laplacian indifference considerations on physical urns suggest that the initial state $p_{v_0}$ of both $\Omega_{DM}^{I}$ and $\Omega_{DM}^{II}$, before any interaction with a cognitive context, is represented by the unit vector
\begin{equation}
|v_0\rangle=\frac{1}{\sqrt{2}}|e_1\rangle+\frac{1}{\sqrt{2}}|e_2\rangle=\frac{1}{\sqrt{2}}(1,1)
\end{equation}
in the ON basis $\{ |e_1\rangle,|e_2\rangle\}$,  A generic state $p_v$ of both $\Omega_{DM}^{I}$ and $\Omega_{DM}^{II}$ is instead represented by the unit vector
\begin{equation} \label{v}
|v \rangle=\rho_R e^{i \theta_{R}}|e_1\rangle+\rho_B e^{i \theta_{B}}|e_2\rangle=(\rho_R e^{i \theta_{R}},\rho_B e^{i \theta_{B}})
\end{equation}
where $\rho_R,\rho_B\ge 0$, $\rho_R^2+\rho_B^2=1$, and $\theta_R,\theta_B\in \Re$.

For every $i\in \{R,B\}$, the (subjective) probability $\mu(E_i,p_{v})$ of drawing a ball of color $i$ in the state $p_v$ of either $\Omega_{DM}^{I}$ or $\Omega_{DM}^{II}$  is then, using (\ref{Bornprobabilitymeasure}) and (\ref{quantumprobability}), given by
\begin{equation}
\mu(E_i,p_v)=\langle v | \hat{P}_{i} | v \rangle=|\langle e_i | v \rangle|^{2}=\rho_{i}^{2}
\end{equation}

Let us now consider the quantum representation of acts. For given utility values $u(0)$ and $u(100)$, the acts $f_1$, $f_2$, $f_3$ and $f_4$ in  Table 1, Sec. \ref{ellsberg}, are respectively represented by the hermitian operators
\begin{eqnarray}
\hat{F}_{1}&=&u(100)\hat{P}_R+u(0)\hat{P}_B \label{f1}\\
\hat{F}_{2}&=&u(100)\hat{P}_R+u(0)\hat{P}_B \label{f2}\\
\hat{F}_{3}&=&u(0)\hat{P}_R+u(100)\hat{P}_B\label{f3} \\
\hat{F}_{4}&=&u(0)\hat{P}_R+u(100)\hat{P}_B \label{f4}
\end{eqnarray}
The EU of $f_1,f_2,f_3$ and $f_4$ in a state $p_{v}$ of both entities $\Omega_{DM}^{I}$ and $\Omega_{DM}^{II}$ is given by
\begin{eqnarray}
W_{v}(f_1)&=&\langle v| \hat{F}_{1}|v\rangle=\rho_{R}^{2}u(100)+\rho_{B}^{2}u(0)=\rho_{R}^{2}u(100)+(1-\rho_{R}^{2})u(0) \label{w1}\\
W_{v}(f_2)&=&\langle v| \hat{F}_{2}|v\rangle=\rho_{R}^{2}u(100)+\rho_{B}^{2}u(0)=\rho_{R}^{2}u(100)+(1-\rho_{R}^{2})u(0)  \label{w2} \\
W_{v}(f_3)&=&\langle v| \hat{F}_{3}|v\rangle=\rho_{R}^{2}u(0)+\rho_{B}^{2}u(100)=\rho_{R}^{2}u(0)+(1-\rho_{R}^{2})u(100) \label{w3}\\
W_{v}(f_4)&=&\langle v| \hat{F}_{4}|v\rangle=\rho_{R}^{2}u(0)+\rho_{B}^{2}u(100)=\rho_{R}^{2}u(0)+(1-\rho_{R}^{2})u(100) \label{w4}
\end{eqnarray}
respectively, where we have used (\ref{v}) and (\ref{f1})--(\ref{f4}).

Coming to the decision process, when a decision-maker is asked to compare acts $f_1$ and $f_2$, the comparison itself, {\it before a decision is taken}, defines a cognitive context, which may change the state of entities $\Omega_{DM}^{I}$ and $\Omega_{DM}^{II}$. Analogously, when a decision-maker is asked to compare acts $f_3$ and $f_4$, the comparison itself, {\it before a decision is taken}, defines a new cognitive context, which may change the state of $\Omega_{DM}^{I}$ and $\Omega_{DM}^{II}$. However, a comparison between $f_1$ and $f_2$ (and also a comparison between $f_3$ and $f_4$) will have different effects on $\Omega_{DM}^{I}$ and $\Omega_{DM}^{II}$. Indeed, since act $f_1$ is ambiguous whereas $f_2$ is unambiguous, a comparison between $f_1$ and $f_2$ will determine a change of $\Omega_{DM}^{I}$ from the state $p_{v_0}$ to a generally different state $p_{v_{12}}$, whereas the same comparison will leave $\Omega_{DM}^{II}$ in the initial state $p_{v_0}$. Analogously, since $f_3$ is ambiguous whereas $f_4$ is unambiguous, a comparison between $f_3$ and $f_4$ will determine a change of $\Omega_{DM}^{I}$ from the state $p_{v_0}$ to a generally different state $p_{v_{34}}$, whereas the same comparison will leave $\Omega_{DM}^{II}$ in the initial state $p_{v_0}$.

Thus, the EUs in (\ref{w2}) and (\ref{w4}) in the state $p_{v_0}$ of $\Omega_{DM}^{II}$ become 
\begin{equation}\label{w2w4}
W_{v_0}(f_2)=W_{v_0}(f_4)=\frac{1}{2} \Big  (u(100)+u(0) \Big )
\end{equation}
which do not depend on the conceptual state of $\Omega_{DM}^{II}$, in agreement with the fact that $f_2$ and $f_4$ are unambiguous acts, while the EUs in (\ref{w1}) and (\ref{w3}) do depend on the final state of $\Omega_{DM}^{I}$, again in agreement with the fact that $f_1$ and $f_3$ are ambiguous acts.

Let us then prove that two {\it ambiguity averse final states} $p_{v_{12}}$ and $p_{v_{34}}$ of $\Omega_{DM}^{I}$ can be found such that the corresponding EUs satisfy the Ellsberg preferences in Sec. \ref{ellsberg}, that is, $W_{v_{12}}(f_1)<W_{v_0}(f_2)$ and $W_{v_{34}}(f_3)<W_{v_0}(f_4)$. Indeed, consider the states $p_{v_{12}}$ and $p_{v_{34}}$ respectively represented, in the canonical ON basis of $\mathbb{C}^{2}$, by the unit vectors
\begin{eqnarray}
|v_{12}\rangle&=&(\sqrt{\alpha},\sqrt{1-\alpha}) \label{v12} \\
|v_{34}\rangle&=&(\sqrt{1-\alpha},-\sqrt{\alpha}) \label{v34} 
\end{eqnarray}
where $0\le \alpha<\frac{1}{2}$. One preliminarily observes that the vectors $|v_{12}\rangle$ and $|v_{34}\rangle$ are orthogonal, that is, $\langle v_{12}| v_{34}\rangle=0$. Moreover, using (\ref{w1})--(\ref{v34}), we get
\begin{eqnarray}
W_{v_{12}}(f_1)&=&\alpha u(100)+(1-\alpha)u(0)<\frac{1}{2}\Big ( u(100)+u(0) \Big )=W_{v_0}(f_2) \\
W_{v_{34}}(f_3)&=&(1-\alpha) u(0)+\alpha u(100)<\frac{1}{2} \Big ( u(100)+u(0) \Big )=W_{v_0}(f_4)
\end{eqnarray}
Hence, the ambiguity averse states $p_{v_{12}}$ and $p_{v_{34}}$ satisfy Ellsberg preferences in the two-urn example within the quantum-theoretic framework. We now intend to further specify $p_{v_{12}}$ and $p_{v_{34}}$, which will allow us to represent the data in Sec. \ref{ellsberg}. This is the aim of Sec. \ref{twourndata}.

\subsection{Data representation in the two-urn example\label{twourndata}}
To represent the data in Sec. \ref{ellsberg}, let us describe the {\it decision between acts $f_1$ and $f_2$} as a measurement with two outcomes that is performed on the entity $\Omega_{DM}^{I}$ in the state $p_{v_{12}}$. In the canonical quantum representation in Hilbert space, this measurement is represented by the spectral family $\{ M, \mathbbmss{1}-M \}$. We assume that the orthogonal projection operator $M$ projects onto the 1-dimensional subspace generated by the unit vector $|m\rangle=(\rho_m e^{i \theta_m}, \tau_m e^{i \phi_m})$, with $\rho_m,\tau_m\ge 0$, $\rho_m^2+\tau_m^2=1$, $\theta_m,\phi_m\in \Re$. Thus, the operator $M$ can be written as
\begin{equation} \label{projM}
M=|m\rangle \langle m|=\left( 
\begin{array}{cc}
\rho_m^2 & \rho_m \tau_m e^{i (\theta_m-\phi_m)} \\
\rho_m \tau_m e^{-i (\theta_m-\phi_m)} & \tau_m^2 
\end{array} \right)
\end{equation}
Analogously, let us describe the {\it decision between acts $f_3$ and $f_4$} as a measurement with two outcomes that is performed on the entity $\Omega_{DM}^{I}$ in the state $p_{v_{34}}$. In the canonical Hilbert space representation, this measurement is represented by the spectral family $\{ N, \mathbbmss{1}-N \}$, where we assume that the orthogonal projection operator $N$ projects onto the 1-dimensional subspace generated by the unit vector $|n\rangle=(\rho_n e^{i \theta_n}, \tau_n e^{i \phi_n})$, with $\rho_n,\tau_n\ge 0$, $\rho_n^2+\tau_n^2=1$, $\theta_n,\phi_n\in \Re$. Thus, the operator $M$ can be written as
\begin{equation} \label{projN}
N=|n\rangle \langle n|=\left( 
\begin{array}{cc}
\rho_n^2 & \rho_n \tau_n e^{i (\theta_n-\phi_n)} \\
\rho_n \tau_n e^{-i (\theta_n-\phi_n)} & \tau_n^2 
\end{array} \right)
\end{equation}
To find a mathematical representation, we have to determine the unit vectors $|v_{12}\rangle$, $|v_{34}\rangle$, $|m\rangle$ and $|n\rangle$ which satisfy the following conditions
\begin{eqnarray}
\langle v_{12}|M|v_{12}\rangle&=&|\langle m|v_{12}\rangle|^2=0.82 \label{exp12}\\
\langle v_{0}|M|v_{0}\rangle&=&|\langle m|v_{0}\rangle|^2=0.50 \label{sym12}\\
\langle m|m\rangle&=&1 \label{norm12} \\
\langle v_{34}|N|v_{34}\rangle&=&|\langle n|v_{34}\rangle|^2=0.84 \label{exp34} \\
\langle v_{0}|N|v_{0}\rangle&=&|\langle n|v_{0}\rangle|^2=0.50 \label{sym34} \\
\langle n|n\rangle&=&1 \label{norm34}
\end{eqnarray}
The system of 6 equations must be satisfied by the parameters $0\le \alpha<\frac{1}{2}$, $\rho_m, \tau_m, \rho_n, \tau_n \ge 0$, $\theta_m, \phi_m, \theta_n,\phi_n \in \Re$. Equations (\ref{exp12}) and (\ref{exp34}) are determined by empirical data, (\ref{norm12}) and (\ref{norm34}) are determined by normalization conditions, while (\ref{sym12}) and (\ref{sym34}) are determined by the fact that decision-makers who are not sensitive to ambiguity should overall be indifferent between $f_1$ and $f_2$, as well as between $f_3$ and $f_4$. Hence, on average, half respondents are expected to prefer $f_1$ ($f_3$) and the other half $f_2$ ($f_4$). To simplify the analysis, let us set $\theta_m=90^{\circ}$, $\theta_n=270^{\circ}$, $\phi_m=\phi_n=0$. Hence, we are left with a system of 6 equations in 5 unknown variables whose solution is
\begin{eqnarray}
\left\{ \begin{array}{lll}    
\alpha&=&0.14815 \\
\rho_m&=&0.21274 \\
\tau_m&=&0.97711 \\
\rho_n&=&0.99155 \\
\tau_n&=&0.12975
 \end{array} \right.
\end{eqnarray}
Hence, the ambiguity averse states $p_{v_{12}}$ and $p_{v_{34}}$ of DM entity $\Omega_{DM}^{I}$ are respectively represented by the unit vectors
\begin{eqnarray}
|v_{12}\rangle&=&(0.38490,0.92296) \\
|v_{34}\rangle&=&(0.92296,-0.38490)
\end{eqnarray}
which in particular determine, through the quantum probability formula (\ref{Bornprobabilitymeasure}), the subjective probability distributions underlying the DM test in Sec. \ref{ellsberg}.

The orthogonal projection operators in (\ref{projM}) and (\ref{projN}) reproducing the preference rates in the same test are instead given by
\begin{eqnarray}
M&=&\left(
\begin{array}{cc}
0.04526 & 0.20787i\\
-0.20787i & 0.95474
\end{array} 
\right) \\
N&=&\left(
\begin{array}{cc}
0.98316 & -0.12865i\\
0.12865i & 0.01684
\end{array} \right)
\end{eqnarray}
The ambiguity averse states reproduce Ellsberg preferences and represent the ambiguity aversion pattern identified in the DM test on the two-urn example, which completes the construction of a quantum mathematical representation for the data. As we can see, ambiguity aversion can be described by means of
{\it genuine quantum structures}, namely, context-dependence, superposition and intrinsically non-deterministic state change, while quantum probabilities represent subjective probabilities.

\section{A general model of hope and fear effects\label{quantumhopefear}}
We apply in this section the quantum-theoretic framework 
exposed
in Secs. \ref{QEUT} and \ref{quantumattitudes} to model hope effects and fear effects in management decisions involving comparison of a risky with an ambiguous option. To this aim, we need to convert the DM tests in Sec. \ref{ambiguityattitudes} into a version of the Ellsberg two-urn example.

We split our analysis into two parts, a {\it gain scenario} (Sec. \ref{gain}) and a {\it loss scenario} (Sec. \ref{loss}).

\subsection{Gain scenario\label{gain}}
We consider the experimental design in Sec. \ref{ambiguityattitudes} and denote by $\bar{p}$ the probability that the value of the financial parameter $\lambda$ in a given investment is {\it above} a benchmark $\lambda_{{\rm benchmark}}$.

We introduce an ambiguous option I with a probability of realizing a gain $G$ which ranges between $\bar{p}-\Delta$ and $\bar{p}+\Delta$, and a risky option with a probability $\bar{p}$ of realizing the gain $G$. This is equivalent to the scenario presented in Table 2. A choice has to be made between acts $f_1$ and $f_2$.

\noindent 
\begin{table} \label{table02}
\begin{center}
\begin{tabular}{|p{0.7cm}|p{2.3cm}|p{2.3cm}||p{2.3cm}|p{2.3cm}|}
\hline
\multicolumn{1}{|c|}{} & \multicolumn{2}{c||}{Option I} & \multicolumn{2}{c|}{Option II} \\
\hline
 Acts & $E_1$: gain & $E_2$: not gain & $E_1$: gain & $E_2$: not gain \\ 
\multicolumn{1}{|c|}{} & \multicolumn{1}{c|}{$p_{1} \in [\bar{p}-\Delta,\bar{p}+\Delta]$}  & \multicolumn{1}{c||}{$p_{2}=1-p_{1}$} &  \multicolumn{1}{c|}{$p_1=\bar{p}$} & \multicolumn{1}{c|}{$p_2=1-\bar{p}$}  \\
\hline
$f_1$ & $G$ & 0 &  &  \\ 
\hline
$f_2$ &  &  & $G$ & 0  \\ 
\hline
\end{tabular}
\end{center}
{\bf Table 2.} Representation of events, payoffs and acts in the gain scenario.
\end{table}

The empirical pattern found in previous studies (see, e.g., \cite{ejt2012,ms2014,vc1999}) and confirmed in \cite{hkk2002} for high probabilities is the following:

(i) most people will prefer $f_2$ to $f_1$ if $\bar{p}$ is high, thus indicating ambiguity aversion and a fear effect;

(ii) most people will prefer $f_1$ to $f_2$ if $\bar{p}$ is low, thus indicating ambiguity seeking and a hope effect.

To reproduce (i) and (ii) in a quantum-theoretic model, we refer to the mathematical formalism in Sec. \ref{quantumellsberg}. More precisely, we introduce a DM entity $\Omega_{DM}^{I}$, corresponding to the ambiguous option, whose initial state $p_{v_0}$ in the absence of any context is represented by the unit vector
\begin{equation} 
|v_0\rangle=\sqrt{\bar{p}}|e_1\rangle+\sqrt{1-\bar{p}}|e_2\rangle=(\sqrt{\bar{p}},\sqrt{1-\bar{p}})
\end{equation}
in the canonical ON basis $\{ |e_1\rangle,|e_2\rangle\}$ of $\mathbb{C}^2$. This initial state will however change under interaction with a context, e.g., the decision-maker pondering between $f_1$ and $f_2$, into a final state $p_v$ represented by the unit vector
\begin{equation} \label{v_hopefears}
|v\rangle=\rho_1e^{i \theta_1}|e_1\rangle+\rho_2 e^{i \theta_2}|e_2\rangle=(\rho_1e^{i \theta_1},\rho_2 e^{i \theta_2})
\end{equation}
where $\rho_1\in [\bar{p}-\Delta,\bar{p}+\Delta]$, $\rho_1^2+\rho_2^2=1$, $\theta_1,\theta_2\in \Re$.

Then, we introduce a DM entity $\Omega_{DM}^{II}$, corresponding to the risky option, whose initial state $p_{v_0}$ in the absence of any context is again represented by the unit vector
\begin{equation} \label{v0_hopefears}
|v_0\rangle=\sqrt{\bar{p}}|e_1\rangle+\sqrt{1-\bar{p}}|e_2\rangle=(\sqrt{\bar{p}},\sqrt{1-\bar{p}})
\end{equation}
This time, however, $p_{v_0}$ is not supposed to change under the interaction with a context, e.g., decision-maker pondering between $f_1$ and $f_2$.

Acts $f_1$ and $f_2$ are instead represented by the hermitian operators
\begin{eqnarray}
\hat{F}_{1}&=&u(G)|e_1\rangle\langle e_1|+u(0)|e_2\rangle\langle e_2| \label{f1_hopefears}\\
\hat{F}_{2}&=&u(G)|e_1\rangle\langle e_1|+u(0)|e_2\rangle\langle e_2| \label{f2_hopefears}
\end{eqnarray}
respectively, where $u(\cdot)$ is the corresponding utility function, such that $u(G)>u(0)$.

We now distinguish between two cases.

{\it Case with high probability $\bar{p}$}. Let us construct a final ambiguity averse state $p_{v_{GH}}$ of DM entity $\Omega_{DM}^{I}$ which reproduces a fear effect, hence such that $W_{v_{GH}}(f_1)<W_{v_0}(f_2)$.

Firstly, the EU of act $f_2$ in the state $p_{v_0}$ of $\Omega_{DM}^{II}$ is, using (\ref{quantumexpectedutility}), (\ref{v0_hopefears}) and (\ref{f2_hopefears}),
\begin{equation}
W_{v_0}(f_2)=\bar{p}u(G)+(1-\bar{p})u(0)
\end{equation}
We choose the final state of $\Omega_{DM}^{I}$ to be the state represented by the unit vector
\begin{equation} \label{vGH_hopefears}
|v_{GH}\rangle=\sqrt{\bar{p}-\alpha}|e_1\rangle+\sqrt{1-\bar{p}+\alpha}|e_2\rangle=(\sqrt{\bar{p}-\alpha},\sqrt{1-\bar{p}+\alpha})
\end{equation}
where $0\le \alpha\le \Delta$. This vector represents an ambiguity averse state. Indeed, the EU of act $f_1$ in the state $p_{v_{GH}}$ of $\Omega_{DM}^{I}$ is, (\ref{quantumexpectedutility}), (\ref{f1_hopefears}) and (\ref{vGH_hopefears}),
\begin{eqnarray}
W_{v_{GH}}(f_1)=(\bar{p}-\alpha)u(G)+(1-\bar{p}+\alpha)u(0)
\nonumber \\
=\bar{p}u(G)+(1-\bar{p})u(0)-\alpha(u(G)-u(0)) <W_{v_0}(f_2)
\end{eqnarray}
Hence, the conceptual state $p_{v_{GH}}$ will generate a fear effect in the gain scenario with high probability $\bar{p}$.

{\it Case with low probability $\bar{p}$}. Let us construct a final ambiguity seeking state $p_{v_{GL}}$ of DM entity $\Omega_{DM}^{I}$ which reproduces a hope effect, hence such that $W_{v_{GL}}(f_1)>W_{v_0}(f_2)$.

We choose the final state of $\Omega_{DM}^{I}$ to be the state represented by the unit vector
\begin{equation} \label{vGL_hopefears}
|v_{GL}\rangle=\sqrt{\bar{p}+\alpha}|e_1\rangle+\sqrt{1-\bar{p}-\alpha}|e_2\rangle=(\sqrt{\bar{p}+\alpha},\sqrt{1-\bar{p}-\alpha})
\end{equation}
where $0\le \alpha\le \Delta$. This vector represents an ambiguity seeking state. Indeed, the EU of act $f_1$ in the state $p_{v_{GL}}$ of $\Omega_{DM}^{I}$ is, using (\ref{quantumexpectedutility}), (\ref{f1_hopefears}) and (\ref{vGL_hopefears}),
\begin{eqnarray}
W_{v_{GL}}(f_1)=(\bar{p}+\alpha)u(G)+(1-\bar{p}-\alpha)u(0)
\nonumber \\
=\bar{p}u(G)+(1-\bar{p})u(0)+\alpha(u(G)-u(0))>W_{v_0}(f_2)
\end{eqnarray}
Hence, the conceptual state $p_{v_{GL}}$ will generate a hope effect in the gain scenario with low probability $\bar{p}$.

\subsection{Loss scenario\label{loss}}
We again consider the experimental design in Sec. \ref{ambiguityattitudes} and proceed as in Sec. \ref{gain}, with obvious changes.

We denote by $\bar{p}$ the probability that the value of the financial parameter $\lambda$ in a given investment is {\it below} a benchmark $\lambda_{{\rm benchmark}}$.

We introduce an ambiguous option I with a probability of realizing a loss $L$ which ranges between $\bar{p}-\Delta$ and $\bar{p}+\Delta$, and a risky option with a probability $\bar{p}$ of realizing the loss $L$. This is equivalent to the scenario presented in Table 3. A choice has been made between acts $f_1$ and $f_2$.

\noindent 
\begin{table} \label{table03}
\begin{center}
\begin{tabular}{|p{0.7cm}|p{2.3cm}|p{2.3cm}||p{2.3cm}|p{2.3cm}|}
\hline
\multicolumn{1}{|c|}{} & \multicolumn{2}{c||}{Option I} & \multicolumn{2}{c|}{Option II} \\
\hline
 Acts & $E_1$: loss & $E_2$: not loss & $E_1$: loss & $E_2$: not loss \\ 
\multicolumn{1}{|c|}{} & \multicolumn{1}{c|}{$p_{1} \in [\bar{p}-\Delta,\bar{p}+\Delta]$}  & \multicolumn{1}{c||}{$p_{2}=1-p_{1}$} &  \multicolumn{1}{c|}{$p_1=\bar{p}$} & \multicolumn{1}{c|}{$p_2=1-\bar{p}$}  \\
\hline
$f_1$ & $L$ & 0 &  &  \\ 
\hline
$f_2$ &  &  & $L$ & 0  \\ 
\hline
\end{tabular}
\end{center}
{\bf Table 3.} Representation of events, payoffs and acts in the loss scenario.
\end{table}

The empirical pattern found in previous studies (see, e.g., \cite{ejt2012,ms2014,vc1999}) and confirmed in \cite{hkk2002} for high probabilities is the following:

(i) most people will prefer $f_1$ to $f_2$ if $\bar{p}$ is high, thus indicating ambiguity seeking and a hope effect;

(ii) most people will prefer $f_2$ to $f_1$ if $\bar{p}$ is low, thus indicating ambiguity aversion and a fear effect.

We again distinguish between two cases and use the symbols and procedures in Sec. \ref{gain}. In particular, (\ref{v0_hopefears})--(\ref{v_hopefears}) still hold and (\ref{f1_hopefears}) and (\ref{f2_hopefears} hold with $L$ in place of $G$ and $u(L)<u(0)$. It is then easy to prove that the ambiguity seeking state of DM entity $\Omega_{DM}^{I}$ generating a hope effect in the loss scenario with high probability $\bar{p}$ is the state $p_{v_{LH}}$ represented by the unit vector
\begin{equation}
|v_{LH}\rangle=\sqrt{\bar{p}-\alpha}|e_1\rangle+\sqrt{1-\bar{p}+\alpha}|e_2\rangle=(\sqrt{\bar{p}-\alpha},\sqrt{1-\bar{p}+\alpha})
\end{equation}
where $0\le\alpha\le \Delta$, while the ambiguity averse state of DM entity $\Omega_{DM}^{I}$ generating a fear effect in the loss scenario with low probability $\bar{p}$ is the state $p_{v_{LL}}$ represented by the unit vector
\begin{equation}
|v_{LL}\rangle=\sqrt{\bar{p}+\alpha}|e_1\rangle+\sqrt{1-\bar{p}-\alpha}|e_2\rangle=(\sqrt{\bar{p}+\alpha},\sqrt{1-\bar{p}-\alpha})
\end{equation}
where again $0\le\alpha\le \Delta$. 

We have thus provided a quantum-theoretic model of both hope and fear effects underlying individual attitudes towards ambiguity in management decisions, like those performed in \cite{vc1999} and \cite{hkk2002}.

\subsection{Data representation\label{managerialdata}}
We elaborate here a Hilbert space representation of the DM test in \cite{hkk2002}, along the lines Secs. \ref{quantumellsberg}, \ref{gain} and \ref{loss}. To this end, we preliminarily note that only high probabilities of gains and losses were considered those studies.

We start by the IRR experiment in Sec. \ref{ambiguityattitudes}, which has the IRR as a parameter of financial performance. Here, we have $IRR_{\rm benchmark}=15\%$. The authors set a probability $\bar{p}=0.68$ and a range $\Delta=0.21$ in the gain scenario, and a probability $\bar{p}=0.65$ and a range $\Delta=0.20$ in the loss scenario. The rate of preference of $f_2$ over $f_1$ was 0.62 in the gain scenario and 0.35 in the loss scenario.

Following the procedure in Sec. \ref{twourndata} and the symbols in Secs. \ref{gain} and \ref{loss}, in a quantum-theoretic representation we need to determine two final states $p_{v_{GH}}$ and $p_{v_{LH}}$, respectively represented in the canonical ON basis of $\mathbb{C}^2$ by the unit vectors
\begin{eqnarray}
|v_{GH}\rangle&=&(\sqrt{0.68-\alpha},\sqrt{0.32+\alpha}) \\
|v_{LH}\rangle&=&(\sqrt{0.65-\alpha},\sqrt{0.35+\alpha})
\end{eqnarray}
$0\le \alpha \le 0.20$, and two 1-dimensional orthogonal projection operators $M=|m\rangle\langle m|$ and $N=|n\rangle\langle n|$ such that
\begin{eqnarray}
\langle v_{GH}|M|v_{GH}\rangle&=&0.62 \\
\langle v_{LH}|N|v_{LH}\rangle&=&0.35
\end{eqnarray}
One can show that a solution is obtained with $\alpha=0.05$, thus
\begin{eqnarray}
|v_{GH}\rangle&=&(0.79373,0.60828) \\
|v_{LH}\rangle&=&(0.77460,	0.63246)
\end{eqnarray}
and
\begin{eqnarray}
M&=&\left(
\begin{array}{cc}
0.96154 & 0.19231 i \\
-0.19231 i & 0.03846
\end{array} 
\right) \\
N&=&\left(
\begin{array}{cc}
0.93733 & -0.24236\\
-0.24236 & 0.06267
\end{array} \right)
\end{eqnarray}

We finally consider the ROI experiment in Sec. \ref{ambiguityattitudes}, where $ROI_{\rm benchmark}=16\%$. The authors set a probability $\bar{p}=0.63$ and a range $\Delta=0.21$ in the gain scenario, and a probability $\bar{p}=0.60$ and a range $\Delta=0.20$ in the loss scenario. The rate of preference of $f_2$ over $f_1$ was 0.66 in the gain scenario and 0.41 in the loss scenario.

Proceeding as above, we need to determine two final states $p_{v_{GH}}$ and $p_{v_{LH}}$, respectively represented by the unit vectors
\begin{eqnarray}
|v_{GH}\rangle&=&(\sqrt{0.63-\alpha},\sqrt{0.37+\alpha}) \\
|v_{LH}\rangle&=&(\sqrt{0.60-\alpha},\sqrt{0.40+\alpha})
\end{eqnarray}
$0\le \alpha \le 0.20$, and two 1-dimensional orthogonal projection operators $M=|m\rangle\langle m|$ and $N=|n\rangle\langle n|$ such that
\begin{eqnarray}
\langle v_{GH}|M|v_{GH}\rangle&=&0.66 \\
\langle v_{LH}|N|v_{LH}\rangle&=&0.41
\end{eqnarray}
One can show that a solution is obtained again with $\alpha=0.05$, thus
\begin{eqnarray}
|v_{GH}\rangle&=&(0.76158,	0.64807) \\
|v_{LH}\rangle&=&(0.74162,	0.67082)
\end{eqnarray}
and
\begin{eqnarray}
M&=&\left(
\begin{array}{cc}
0.99312 & 0.08215 \\
0.08215 & 0.00679
\end{array} 
\right) \\
N&=&\left(
\begin{array}{cc}
0.12500 & 0.33072 i\\
-0.33072 i & 0.12500
\end{array} \right)
\end{eqnarray}
This completes the construction of a quantum representation of the experimental study presented in \cite{hkk2002}. As we can see, the influence of psychological factors, like hopes and fears, on decision inherent investments, can be incorporated within a quantum-theoretic framework, while aversion to ambiguity and ambiguity seeking behaviour arise as effects due to genuine quantum structures, like context-dependence, superposition and intrinsically non-deterministic state change. The presence of quantum structures explains the departure of rationality, in either direction (aversion, seeking) and also the switch between a direction and another.

\section{Conclusions\label{conclusions}}
In this paper, we have put forward a theoretical framework that uses the Hilbert space formalism of quantum theory to model the deviations from subjective EUT observed in concrete human decisions. In this framework, ambiguity aversion and ambiguity seeking behaviour are described as effects due to genuine quantum structures, namely, context-dependence, superposition and non-deterministic change of state. The quantum-theoretic framework enables in this way representation of various sets of empirical data in simple bets on urns, as well as in management decisions.

We conclude this paper with a comment that is relevant, in our opinion, to better grasp the content of our findings.

One may object that the quantum models in Secs. \ref{QEUT}, \ref{quantumattitudes} and \ref{quantumhopefear} are ``ad hoc'', in the sense that they require introduction of new parameters,  which would describe empirical data, without necessarily explaining them. In particular, one may wonder why we have chosen to use a Hilbert space over complex numbers, whereas a Hilbert space over real numbers would have been sufficient to reproduce ambiguity seeking and ambiguity aversion attitudes, which has already been proved (see, e.g., \cite{lm2009,and2017}).

It is then worth emphasizing that our main aim in this paper was exactly developing a ``unitary theory of human DM'', rather than studying specific DM situations. In the investigation of specific situations in which an interaction occurs of a decision-maker with a DM entity, we firstly look for a theory of DM, i.e. the theoretical framework elaborated in Sec. \ref{QEUT}, truly describing ``the reality of the cognitive realm to which a DM entity belongs'' and, additionally, ``how human minds can interact with the latter so that a decision occurs''. In this sense, each time we elaborate a model for a given cognitive effect, e.g., the models in Secs. \ref{quantumattitudes} and \ref{quantumhopefear}, it is always our preoccupation to also make sure that: (i) the model is derived following the logic that governs our theory of DM -- in our case, quantum theory in complex Hilbert space, and (ii) whatever other tests were to be performed by a human mind interacting with the DM entity, also the data of these hypothetical additional tests could be modelled following our theory of DM. Clearly, the requirement that ``all possible tests and data'' have to be modelled in an equivalent way poses severe constraints to our theory, and it is not a priori evident that this would always be possible. However, we believe that the fundamental idea underlying our methodology, namely, looking upon a decision as an interaction of a human mind with a DM entity in a specific state, equips the theory of exactly those degrees of freedom that are needed to model ``all possible data from all possible tests''. Being ``theory-based'', rather than ``data-based'', the models following from our theory of DM are not``ad hoc'', though the behavioural meaning of the parameters appearing there is not trivial to interpret at this stage of the research. 
 
Of course, new DM tests have to be performed to check whether the quantum-theoretic framework and ensuing models work in general, or whether we instead need more complex Kolmogorovian or more general non-Hilbertian structures in a unified theory of human decisions.

\appendix

\section{The mathematics of quantum theory in Hilbert space}\label{quantummathematics}
We present here the fundamental terminology, definitions and axioms of the formalism of quantum theory in Hilbert space that are needed to grasp the content of this paper. More detailed introductions to the Hilbert space formalism for cognitive and social scientists are provided in, e.g., \cite{bb2012} and \cite{hk2013}.

When the formalism of quantum theory is applied for modelling purposes, each entity that is considered is associated with a {\it Hilbert space} $\mathscr H$, that is, a finite-dimensional vector space over the field ${\mathbb C}$ of {\it complex numbers}, equipped with a {\it scalar product} $\langle \cdot |  \cdot \rangle$ that maps two vectors $\langle u|$ and $|v\rangle$ onto a complex number $\langle u|v\rangle$. We denote vectors by using the ``bra-ket notation'' that is usual in modern manuals on quantum theory.

Vectors can be {\it kets}, denoted by $\left| u \right\rangle $, $\left| v \right\rangle$, or {\it bras}, denoted by $\left\langle u \right|$, $\left\langle v \right|$. The scalar product between the ket-vectors $|u\rangle$ and $|v\rangle$, or the bra-vectors $\langle u|$ and $\langle v|$, is obtained by juxtaposing the bra-vector $\langle u|$ and the ket-vector $|v\rangle$, and $\langle u|v\rangle$ is also called a {\it bra-ket}. The scalar product $\langle \cdot |  \cdot \rangle$ satisfies the following properties: for every $|u\rangle, |v\rangle, |w\rangle \in \mathscr H$ and $a,b \in {\mathbb C}$,

(i) $\langle u |  u \rangle \ge 0$;

(ii) $\langle u |  v \rangle=\langle v |  u \rangle^{*}$;

(iii) $\langle u |(a|v\rangle+b|w\rangle)=a\langle u |  v \rangle+b \langle u |  w \rangle $.

The term $\langle v |  u \rangle^{*}$ in (ii) denotes the {\it complex conjugate} of $\langle u |  v \rangle$, while the sum vector $a|v\rangle+b|w\rangle$ is called the {\it linear combination} of vectors $|v\rangle$ and $|w\rangle$.

From (ii) and (iii) follows that the scalar product $\langle \cdot |  \cdot \rangle$ is linear in the ket and anti-linear in the bra, that is, $(a\langle u|+b\langle v|)|w\rangle=a^{*}\langle u | w\rangle+b^{*}\langle v|w \rangle$.

We remind that the {\it absolute value} $|z|$ of a complex number $z\in {\mathbb C}$ is defined as the square root of the product between $z$ and its complex conjugate $z^{*}$, that is, $|z|=\sqrt{z^{*}z}$. In addition, a complex number $z$ can either be decomposed into its {\it cartesian form} $z=x+iy$ (where $i$ is the {\it imaginary unit}, or into its {\it polar form} $z=|z|e^{i\theta}=|z|(\cos\theta+i\sin\theta)$.  Hence, we have $|\langle u| v\rangle|=\sqrt{\langle u|v\rangle\langle v|u\rangle}$. We then define the {\it length} of a ket-(bra-)vector $|u\rangle$ ($\langle u|$) as $|| |u\rangle ||=||\langle u |||=\sqrt{\langle u |u\rangle}$. A vector of unitary length is called a {\it unit vector}. We say that two ket-vectors $|u\rangle$ and $|v\rangle$ are {\it orthogonal}, and write $|u\rangle \perp |v\rangle$, if $\langle u|v\rangle=0$.

We have now introduced the necessary mathematics to state the first modelling rule of quantum theory, as follows.

\medskip
\noindent{\it First quantum modelling rule.} A state $p_u$ of an entity modelled by quantum theory is represented by a ket-vector $|u\rangle$ with length 1, that is, $\langle u|u\rangle=1$.

\medskip
\noindent
An {\it orthogonal projection operator} $\hat{P}$ is an operator on the Hilbert space $\mathscr H$, that is, a mapping ${\hat P}: {\mathscr H} \rightarrow {\mathscr H}, |u\rangle \mapsto {\hat{P}}|u\rangle$, such that the following properties are satisfied: for every $|u\rangle, |v\rangle \in {\mathscr H}$ and $a, b \in {\mathbb C}$, 

(i) $\hat{P}(a|u\rangle+b|v\rangle)=a \hat{P} |u\rangle+b \hat{P}|v\rangle$ ({\it linearity});

(ii) $\langle u|\hat{P}|v\rangle=\langle v|\hat{P}|u\rangle^{*}$ ({\it hermiticity});

(iii) $\hat{P}^2=\hat{P}$ ({\it idempotency}).

The identity operator $\mathbbmss{1}$ maps each vector onto itself and is a trivial orthogonal projection operator. We say that two orthogonal projection operators $\hat{P}_i$ and $\hat{P}_j$ are {\it orthogonal operators} if each vector contained in the range $\hat{P}_i({\mathscr H})$ is orthogonal to each vector contained in the range $\hat{P}_j({\mathscr H})$, and we write $\hat{P}_i \perp \hat{P}_j$, in this case. The orthogonality of the orthogonal projection operators $\hat{P}_i$ and $\hat{P}_j$ can also be expressed by $\hat{P}_i\hat{P}_j=0$, where $0$ is the null operator. A set of orthogonal projection operators $\{\hat{P}_i \ | \ i\in \{1,\ldots,n \}\}$ is called a {\it spectral family} if all $\hat{P}_i$s are mutually orthogonal, that is, $\hat{P}_i \perp \hat{P}_j$ for every $i \not= j$, and their sum is the identity operator, that is, $\sum_{i=1}^n \hat{P}_i=\mathbbmss{1}$. A spectral family $\{\hat{P}_i \ | \ i\in \{1,\ldots,n \}\}$ identifies an {\it Hermitian operator} $\hat{O}=\sum_{i=1}^{n}o_i \hat{P}_i$, where $o_i \in \Re$ is called an {\it eigenvalue} of $\hat{O}$, that is, $o_i$ is a solution of the equation $\hat{O}|o\rangle=o|o\rangle$ -- the non-null vectors satisfying this equation are called the {\it eigenvectors} of $\hat{O}$.

The above definitions give us the necessary mathematics to state the second and third quantum modelling rules, as follows.

\medskip
\noindent
{\it Second quantum modelling rule.} A measurable quantity $O$ of an entity modelled by quantum theory is represented by a spectral family $\{\hat{P}_i \ | \ i \in \{ 1, \ldots, n \}\}$ or, equivalently, by an Hermitian operator $\hat{O}=\sum_{i=1}^{n}o_i \hat{P}_i$, where the eigenvales $\{o_1, \ldots, o_n\}$ are the only possible values that can be obtained in a measurement of $O$.

\medskip
\noindent
{\it Third quantum modelling rule.} The probability of obtaining the value $o_i$, $i \in \{ 1, \ldots, n \}$, in a measurement of the quantity $O$, represented by the spectral family $\{\hat{P}_i \ | \ i \in \{ 1, \ldots, n \}\}$, on an entity modelled by quantum theory in a state $p_u$, represented by the unit vector $|u\rangle$, is given by $\langle u| \hat{P}_{i}|u\rangle=||\hat{P}_i |u\rangle||^{2}$. This formula is called the {\it Born rule} in quantum theory. Moreover, if the value $o_i$ is actually obtained in the measurement and the measurement is {\it ideal}, then the initial state $p_u$ of the entity is changed into a state $p_{u_i}$ represented by the vector
\begin{equation}
|u_i\rangle=\frac{\hat{P}_i|u\rangle}{||\hat{P}_i|u\rangle||} \nonumber
\end{equation}
This change of state is called {\it reduction}, or {\it collapse}, in quantum theory.

\medskip
\noindent
Let us now come to the formalization of quantum probability. A major structural difference between classical probability theory, which satisfies the axioms of Kolmogorov, and quantum probability theory, which is non-Kolmo\-gorovian, relies in the fact that the former is defined on a Boolean $\sigma$-algebra of events (see Sec. \ref{SEUT}), whilst the latter is defined on a more general algebraic structure. More specifically, let us denote by ${\mathscr L}({\mathscr H})$ the set of all orthogonal projection operators over the ($n$-dimensional, for the sake of simplicity) complex Hilbert space  ${\mathscr H}$. The set ${\mathscr L}({\mathscr H})$ has the algebraic properties of a {\it complete orthocomplemented lattice}, but ${\mathscr L}({\mathscr H})$  is generally not distributive, hence ${\mathscr L}({\mathscr H})$ does not form a $\sigma$-algebra. Then, a {\it generalized probability measure} over  ${\mathscr L}({\mathscr H})$ is a mapping $\mu: \hat{P} \in {\mathscr L}({\mathscr H}) \longmapsto \mu(\hat{P}) \in [0,1]$, such that $\mu(\mathbbmss{1})=1$, and $\mu(\sum_{i=1}^{n} \hat{P}_i)=\sum_{i=1}^{n}\mu(\hat{P}_i)$, for any countable sequence $\{ \hat{P}_i \in {\mathscr L}({\mathscr H}) \ | \  i\in \{1,\ldots, n \} \}$ of mutually orthogonal projection operators. The elements of ${\mathscr L}({\mathscr H})$ are said to represent {\it quantum events}, in this framework. Referring to the quantum modelling rules above, the event ``a measurement of the quantity $O$ gives the value $o_i$'' is represented by the orthogonal projection operator $\hat{P}_i$ in quantum theory.

Born's rule establishes a connection between states and generalized probability measures, as follows. 

Given a state $p_u$ of an entity modelled by quantum theory, represented by the unit vector $|u\rangle \in \mathscr H$, it is possible to associate  $|u\rangle$ with a generalized probability measure $\mu_{u}$ over ${\mathscr L}({\mathscr H})$, such that, for every $\hat{P} \in {\mathscr L}({\mathscr H})$, $\mu_{u}(\hat{P})=\langle u |\hat{P}|u\rangle$. The generalized probability measure $\mu_{u}$ is a {\it quantum probability measure} over ${\mathscr L}({\mathscr H})$.


\end{document}